# A Multiscale Graph Convolutional Network for Change Detection in Homogeneous and Heterogeneous Remote Sensing Images

Junzheng Wu, Biao Li, Yao Qin, Weiping Ni, Han Zhang and Yuli Sun

*Abstract*—Change detection (CD) in remote sensing images has been an ever-expanding area of research. To date, although many methods have been proposed using various techniques, accurately identifying changes is still a great challenge, especially in the high resolution or heterogeneous situations, due to the difficulties in effectively modeling the features from ground objects with different patterns. In this paper, a novel CD method based on the graph convolutional network (GCN) and multiscale object-based technique is proposed for both homogeneous and heterogeneous images. First, the object-wise high level features are obtained through a pre-trained U-net and the multiscale segmentations. Treating each parcel as a node, the graph representations can be formed and then, fed into the proposed multiscale graph convolutional network with each channel corresponding to one scale. The multiscale GCN propagates the label information from a small number of labeled nodes to the other ones which are unlabeled. Further, to comprehensively incorporate the information from the output channels of multiscale GCN, a fusion strategy is designed using the father-child relationships between scales. Extensive Experiments on optical, SAR and heterogeneous optical/SAR data sets demonstrate that the proposed method outperforms some state-of the-art methods in both qualitative and quantitative evaluations. Besides, the Influences of some factors are also discussed.

## I. INTRODUCTION

CHANGE detection (CD) that aims at identifying the changes of regions or phenomenon in the same geographical area at different times has been an attractive research topic [1]. It has been extensively applied to various fields, such as resources investigation [2], urban growth monitoring [3] and disaster assessment [4].

Up to now, optical and synthetic aperture radar (SAR) images have been two of the most common types of remote sensing (RS) data in CD tasks [5]. To be specific, the great majority of remote sensing images are acquired from optical sensors and they can represent abundant information of land cover (such as texture, structure and color). Therefore, CD with optical images has been of interest [6] for a long time and is relatively mature. On the other hand, the properties that SAR images can be acquired in all-weather and all-time conditions [7] make them reasonably potential in some applications, and literally, CD with SAR images has also been an active scope for researchers. CD with images collected by the same kind of sensors, e.g., SAR or optical sensor, can refer to the homogeneous CD which is the main stream in the current CD field. However, CD with heterogeneous images collected by different types of sensors has great practical significance in some emergency situations (e.g., earthquake or flood) where the rapid mapping of damages is needed. Frequently, only the pre-event optical image can be obtained from the archived data, whereas maybe only the post-event SAR image can be available due to the adverse atmospheric conditions [8]. Consequently, the heterogeneous CD has drawn increasing attention. Generally, heterogeneous CD is particularly challenging because of the distinct feature representation in images acquired by different sensors.

Numerous works have been devoted to both homogeneous and heterogeneous CD tasks in the passing several decades. In the case of homogeneous images, many methods are based on the difference between spectral bands or between intensities. For instance, differencing, change vector analysis (CVA) [9], multivariate alteration detection (MAD) [10] and iteratively reweighted MAD (IR-MAD) [11] are frequently used for optical images while log-ratio (LR) [12] and mean-ratio (MR) [13] are popular in SAR CD tasks. Inspired by the above strategies, plenty of methods have been proposed to generate more reliable difference images (DI). Wu et al [14] proposed a slow feature analysis (SFA) method which minimized the feature variation of unchanged pixels, and meanwhile, changed pixels could be highlighted and separated for multispectral images. Saha et al [15] developed an extended CVA in unsupervised schema, namely deep change vector analysis (DCVA), to obtain robust change vectors that model the spatial

J. Wu, B. Li and Y. Sun are with the Key Laboratory of ATR, College of Electronic Science, National University of Defense Technology, Changsha, 410073, China (e-mail: wujunzheng@nint.ac.cn; libiao@mail.nudt.edu.cn; sunyili19@mail.nudt.edu.cn ).

context information for very high resolution (VHR) images. Lv et al [16] designed an object-oriented key point vector distance to measure the change magnitude between VHR images. Wang et al [17] constructed the relationships among pixels and their coupling neighbours using hypergraphs which can capture both change level of pixels and local consistency, thus enabling to generate DI with a good separability for SAR images. As CD can be treated as classification problems, many researchers have resorted to classification and statistical techniques. Some of them categorized the feature extracted from the input image pairs into changed/unchanged classes by taking advantage of supervised classifiers, e.g., support vector machine (SVM) [18], extreme learning machine (ELM) [19] and random forest (RF) [20]. Some of them made the assumption that the image intensities or DI obey a specific statistical distribution [21]-[23]. For example, Zanetti et al [24] modeled the magnitude of the DI using a Rayleigh-Rice mixture density, via the novel parameter estimation, the mixture model outperformed several empirical models for CD with optical images.

Despite fewer comparing with those for homogeneous CD, quite a few methods have emerged in heterogeneous CD tasks. For instance, Mercier et al [25] transformed one of the two images to the other one using the copula theory, which led to obtain comparable characteristics, then, the Kullkack–Leibler (KL) distance was employed to calculate the change indices. Ferraris et al [26] used the coupled dictionary learning framework to model the two heterogeneous images which can be represented by a sparse linear combination of atoms belonging to a pair of coupled over-complete dictionaries learnt from the two images. Wan et al [27] combines the cooperative multi-temporal segmentation and hierarchical compound classification strategy to overcome the error propagation in classification-based methods. Moreover, multivariate statistical model [28], homogeneous pixel transformation [29], image regression [30] and theory of graph [31] were also developed to learn the latent relationships between heterogeneous images.

Recently, deep learning techniques have demonstrated remarkable performance in image processing field due to their capabilities of automatically obtaining abstract high level representations by gradually aggregating the low-level feature, by which the complicated feature engineering can be avoided [32], and without doubt, various deep learning methods have been employed in CD tasks, including convolutional neural network (CNN), auto-encoder (AE), recurrent neural network (RNN) and deep belief network (DBN). For example, Lyu et al [33] adopted RNN to learn transferable change-rules between multi-temporal homogeneous images. Liu et al [34] proposed a stacked Fisher auto-encoder to extract layer-wise feature which was more discriminative in SAR CD. Chen et al [35] proposed a Siamese convolutional multiple-layers recurrent neural network to extract spatial–spectral feature from homogeneous or heterogeneous image patches. Among these deep learning methods, CNN has drawn intensive attention and has been the most popularly used as the backbone or feature extractors in CD tasks [36]. The CNN-based CD methods will be discussed in more detail in Section II.A.

Although the aforementioned methods have achieved promising performances in some cases, they generally have at least one of the following weaknesses:

(1) Hand-crafted features are needed to extract within most of the difference based and classification based methods, which require much domain-specific knowledge and may be affected by noise and atmospheric conditions. (2) The deep learning based methods rely heavily on large amounts of annotated training samples, which are typically generated by manually labeling or pre-classification, the former is time-consuming and tedious, while the latter requires highly reliable classification results which can rarely be guaranteed in practice. (3) Many methods are designed for only one type of data, consequently, the performance is unsatisfactory when transferred to other types of data. (4) The current state-of-the-art CNN-based methods cannot fully capture the geometric variations of different object regions because the convolutions are only conducted on the regular rectangle regions.

Aiming at addressing or alleviating these shortcomings, a novel method is proposed for CD combining the multiscale object-based technique and the graph convolutional network (GCN) in this paper. First, a pre-trained U-net combining with the multiscale segmentation technique is employed to extract features of the input images. Instead of using high level features for pixels/patches or hand-crafted features for ground objects with varied sizes and shapes, here, the high level features and the comprehensive contextual information can be incorporated. On the basis of treating each object as a graph node with features extracted previously, the multiscale GCN can be constructed and a fusion strategy is designed for the outputs of the different GCN channels based on the relationships between the segmented scales. Finally, a binary change map can be produced owing to the property that GCN can propagate information form a small amount of labeled nodes to the unlabeled ones.

The main contributions of this paper are as follows.

1) We propose a novel multiscale graph convolutional network (MSGCN) for CD which uses multi-channel graph convolutional layers to process the extracted high level features. Moreover, a multiscale decision fusion strategy is designed to make full use of the comprehensive information of these channels, which improves the detection accuracy.

2) The designed feature extractor which adopts a pre-trained U-net integrating with the multiscale segmentation can automatically obtain object-wise high level features to construct the nodes of graphs.

3) The remarkable experimental results have demonstrated that the proposed method can be effectively applied to both homogeneous and heterogeneous high resolution RS images with small amounts of labeled samples, making it practical in various CD applications.

The rest of this article is organized as follows. We discuss the related works in Section II. Section III illustrates the proposed method in detail. Section IV provides the details about the data, experimental results and discussions. Finally, the conclusion is presented in Section V.

## II. RELATED WORKS

In this section, we provide a brief review of CNN-based CD methods and GCN, as they are related to this article.

### A. CNN-based Change Detection

Recently, various CNN-based CD methods have been proposed due to the capabilities of automatically extracting high level semantic features that can avoid the hand-crafted feature design. For instance, Nemoto et al [37] intuitively utilized CNN to classify the images acquired at two periods, respectively, and then, the classifications are compared to obtain change information. Lim et al [38] designed three encoder-decoder structured CNNs to yield change maps using Google Earth images. A fully convolutional Siamese auto-encoder method for CD in UAV images was presented in [39], which can reduce the number of labeled samples required to achieve competitive results. Liu et al [40] used a pre-trained CNN-based U-net architecture and designed a new loss function to achieve transfer learning for CD tasks among different data sets. Similarly, Zhang and Shi [41] used a CNN to learn the deep features and then used a CNN-based transfer learning framework to compose a two channel network with shared weight to generate a multiscale and multi-depth feature difference map for CD. Several generative adversarial network (GAN) architectures based on CNN units have been also exploited for CD [42-44].

Although the existing CNN-based methods have validated the availabilities in many cases, some limitations still exist. To be specific, ground objects in RS images commonly appear with multiscale and various shape, whereas the CNN models only conduct the convolution on the regular rectangular regions. In other works, the shape information of objects cannot be completely captured by CNN. Besides, the weights of each convolution kernel are identical when convolving all patches [45]. As a result, the boundaries between changed and unchanged classes may be lost.

### B. Graph Convolutional Network

The notion of graph neural network (GNN) was initially outlined in [46] and further elaborated in [47-49]. Wu *et al* have made a comprehensive survey on GNN in [50]. As mentioned above, CNN often fails to analyze spatial vector data because of the regularity requirements for data structures. To handle the irregular data such as graph, new generalizations and definitions of the convolution operations have been rapidly developed over the past few years, which can be the umbrella of GCN. For instance, Bruna et al [51] used a spectral convolution to define a multilayer neural network model, which is similar to the classical CNN. Sandryhaila and Mouraj [52] attempted to redefine the convolution as a polynomial function of the graph adjacency matrix. Kipf and Welling [53] proposed a fast approximation localized convolution and designed a simple layer-wise propagation rule for semi-supervised classification, which makes the GCN model able to encode both graph structure and node features. Since then, increasing extensions and improvements have emerged, such as FastGCN [54] and graph convolutional auto-encoder using Laplacian smoothing and sharpening (GALA) [55].

With the capability of modeling the irregular data structures, GCN has been widely applied to various vision tasks, such as semantic segmentation [56], specific object detection [57] and hyperspectral RS image classification [45]. To the best of our knowledge, GCN has been deployed for CD in only one prior work [58]. However, in the work of [58], several hand-crafted features (mean, maximum, and minimum spectral values and area) of the objects under multiscale are simply concatenated to be the features of nodes, and thus, the distinction among objects cannot be fully reflected. To be specific, if two objects under a fine scale are both within the areas of one object under a coarser scale, these two objects would share a majority of the concatenated features, thus, the discrimination may be weaken. Besides, despite multiscale hand-crafted features are extracted, only the objects under the finest scale are treated as nodes to construct the graph model and fed to GCN, which means that multiscale information has not been adequately exploited. To cope with these issues, we proposed a multiscale GCN that uses object-wise features extracted by a pre-trained U-net integrating with the multiscale segmentation as the features of nodes and fuses outputs of the multi-channel networks for CD. As a result, accurate node embedding and enrich multiscale information can be cooperated, which can facilitate the promotion of detection performance.

## III. PROPOSED METHOD

In this section, we elaborate on the details of the proposed method, as shown in Fig.1. Firstly, the two input images are stacked as one image and then fed into the pre-trained feature extracting network and the multiscale segmented module, simultaneously. After that, for each segmented scale, one graph is constructed using the object-wise features of nodes obtained from the extracted high level features combining with the objects under the segmented scale. Then, the multiscale GCN layers are conducted on these graphs, which can cluster the objects potentially belonging to the same class (changed/unchanged) together in the embedding space. Finally, the CD result is produced through fusing the outputs of the multiscale GCN by the designed fusion module.

### A. Multiscale Segmentation

Object-based image analysis uses a homogeneous region (namely object) as the basic processing unit which contains sufficient spatial information. It always starts with object segmentation, thus, creating representative objects through segmentation approaches is crucial for feature extraction. In this paper, fractal net evolution approach (FNEA) [59] which has been demonstrated to be effective [60] in object-based RS image analysis is adopted for the segmentation task. The objects are obtained by merging regions according to an optimization function, which requires the heterogeneity of the merged object in terms of spectral and shape properties to be lower than the user-defined threshold. Compared with other segmentation approaches such as SLIC and mean-shift, FNEA is adopted for CD in this paper based on the following advantages. First, the hierarchical segmentation strategy allows the objects with different sizes to be fully extracted by simply tuning the scale parameter. Second, not only spectral properties but also geometric information are taken into account during segmentation, thus, objects with various shapes can be extracted with relatively high accuracy [27]. In addition, an object under a coarse segmented scale can be obtained by merging several objects under a finer one keeping other parameters are invariant. It means that the father-child relationships exist between segmented scales, and these relationships can be exploited to fuse the multiscale information.

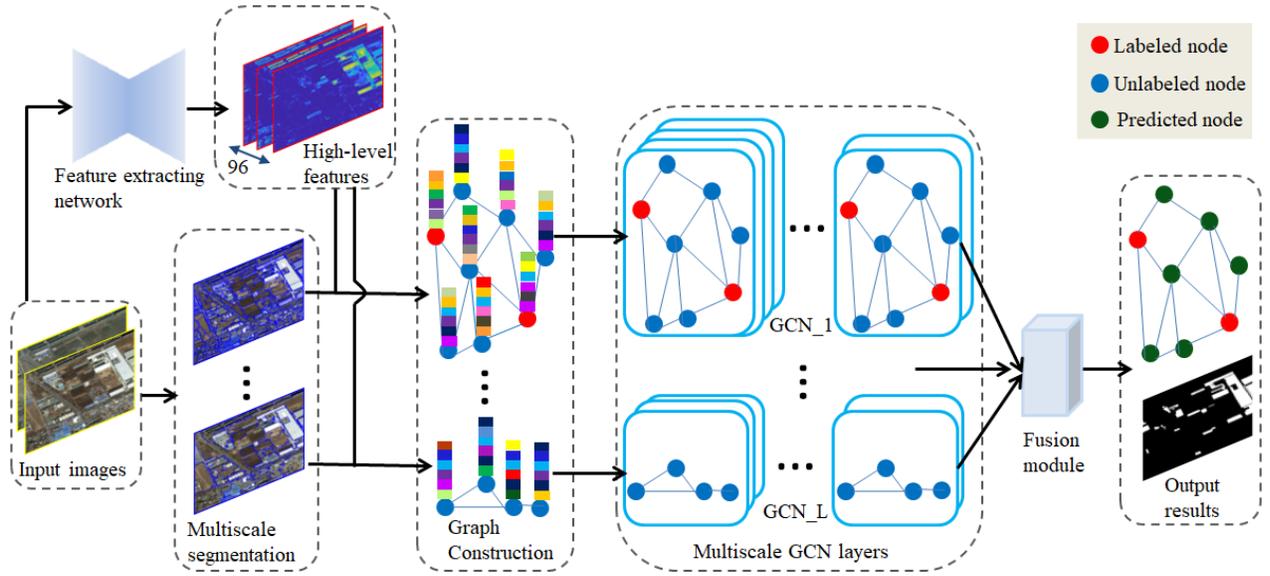

Fig.1. DIs generated by (a) IRMAD(LR for SAR), (b) ITPCA(MR for SAR), (c) PCVA(ITPCA for SAR), (d) OCVA(OMR for SAR), (e) MVKL(OCN for SAR) and (f) proposed approach.

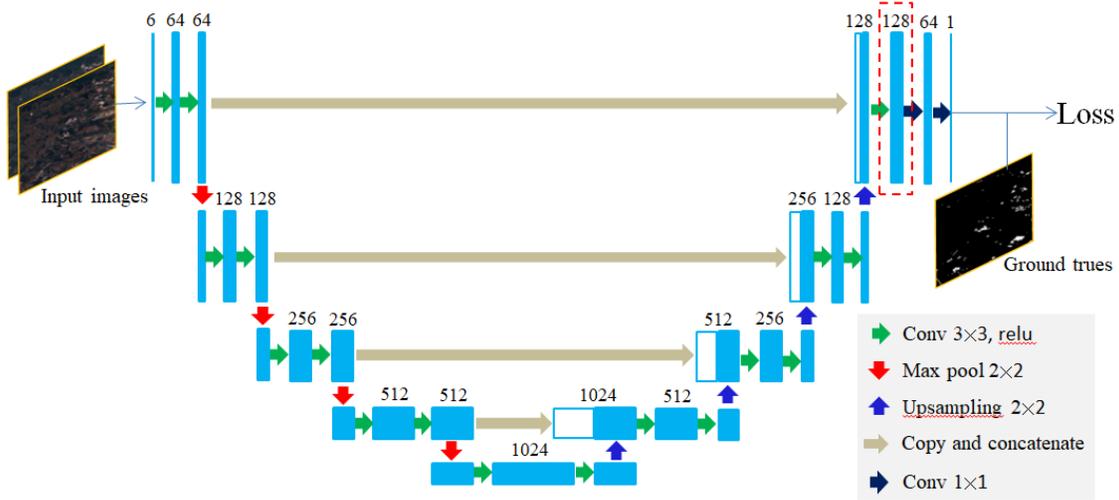

Fig.2. DIs generated by (a) IRMAD(LR for SAR), (b) ITPCA(MR for SAR), (c) PCVA(ITPCA for SAR)

### B. Feature Extracting Network

In many vision tasks, feature extraction and selection is a complex but rather pivotal step that requires professional knowledge and experience. Considering that the improved version of fully convolutional neural network (FCN) called U-net has been effectively exploited to some CD works [61-62], meanwhile, several open RS datasets for CD tasks have been available to train the networks, such as the ONERA Satellite Change Detection (OSCD) dataset [63] and the SZTAKI AirChange Benchmark set [64], we adopted and modified the original U-net structure to extract pixel-wise high level features, which can avoid the manual designs for feature extraction and incomprehensively modeling the features.

As shown in Fig.2, the network is trained in the end-to-end manner with free-size input image. The U-net structure consists three parts that involve a contracting (or encoding) path, bottleneck, and expanding (or decoding) paths. The encoding path which is a typical convolutional neural network structure comprises four sets of convolution operations and four down-sampling operations, among which each convolution set adopts two consecutive 3×3 convolution operations. Each set of convolution operations is dimensioned by a down-sampling operation and the Relu function is used as the activation function. Between the contracting and expanding paths, the bottleneck is built from two simple convolution layers with the kernel size of 3×3. The decoding path is almost symmetric with the decoding path, and it comprises four sets of transposed convolution operations and four up-sampling operations, among which each convolution set uses the concatenation with the corresponding cropped feature maps from contracting path to help the decoder better repair target details. The kernel size of the convolution layers in decoding path is also 3×3. In addition, the final two layers of the network uses two 1×1 convolutional layers that gradually convert the feature maps into one map that indicate the probabilities of change for each pixel. The cross entropy function is adopted as the loss function with the following formula:

$$loss = -\sum_{n}\left[ y_n \ln a_n + (1-y_n)\ln(1-a_n) \right] \quad (1)$$

Where $y_n$ is the real value of the sample, and $a_n$ is the actual output result.

In this paper, we pre-trained the U-net with the openly free OSCD dataset [63]. The input images are clipped into patches with the size of 112×112 in the pre-training process. When the U-net has been trained, it can be used as the feature extracting network as the module shown in Fig.1, and the input images can be free-size without any adjustment. It is worth noting that the 128 feature maps (marked by the red box in Fig.2) of the last convolution operations set are treated as the extracting results of the feature extracting network and fed to the following graph construction.

### C. Graph Construction

Supposing that the stacked input image $I_S$ is segmented by the FNEA approach under $L$ scale parameters ($S_1, \cdots, S_L$, ranging from fine to coarse) with the same of other segmented parameters, respectively. Then, $L$ object sets can be obtained: $\Omega_1 = \{P_{1,1}, \cdots, P_{1,N_1}\}, \cdots, \Omega_L = \{P_{L,1}, \cdots, P_{L,N_L}\}$, where $P_{l,k}$ denotes the $k$th object under the scale parameter $S_l, (l=1,2,\cdots,L)$, and $N_1, \cdots, N_L$ denote the numbers of object corresponding to the scale parameters $S_1, \cdots, S_L$, respectively, and naturally, $N_1 > N_2 > \cdots > N_L$. Treating each object as one node, $L$ undirected affinity graphs $G_1 = \{V_1, E_1\}, \cdots, G_L = \{V_L, E_L\}$ can be constructed, where $v_{l,i} \in V_l$ denotes each object region, and $e_{l,ij} = (v_{l,i}, v_{l,j}) \in E_l$ represents the spatial relationships between two arbitrary objects under the scale parameter $S_l, (l=1,2,\cdots,L)$.

In this paper, we generate the graph-based representation by building features for each node and representing the edges between nodes using an adjacency matrix under each scale parameter.

Through the pre-trained feature extracting network, 128 high level feature maps $\mathbf{M}_1, \cdots, \mathbf{M}_{128}$ can be extracted, and each of them has the same size with the input image ($H \times W$). To build the object-wise feature vector $\mathbf{F}_{l,i} \in \mathbf{R}^{1\times128}$ representing the $i$th node under the scale parameter $S_l$, we combine the above feature maps which represent high level semantic information and the segmentation result $\Omega_l = \{P_{l,1}, \cdots, P_{l,N_l}\}$ that contain abundantly spatial information to characterize the node. $\mathbf{F}_{l,i}$ is formulated by:

$$\mathbf{F}_{l,i}(n) = \sum_{\{(j,k)|I_s(j,k)\in P_{l,i}\}} M_n(j,k) / \#(P_{l,i}), n=1,2,\cdots,128 \quad (2)$$

where $\#(P_{l,i})$ represents the number of pixels in $P_{l,i}$.

Given the $l$th segmentation set $\Omega_l = \{P_{l,1}, \cdots, P_{l,N_l}\}$, we aim to obtain an adjacency matrix $\mathbf{A}_l \in \mathbf{R}^{N_l \times N_l}$ which indicates the interaction of each pair of nodes in $G_l$. The elements of $\mathbf{A}_l$ can be calculated as:

$$\mathbf{A}_{l,ij} = \begin{cases} e^{-d(i,j)} \cdot e^{-\gamma \|F_{l,i}-F_{l,j}\|}, P_{l,i} \in Nei(P_{l,j}) | P_{l,j} \in Nei(P_{l,i}) \\ 0, \quad otherwise \end{cases} \quad (3)$$

Where $Nei(P_{l,i})$ is the set of neighbors of the object $P_{l,i}$, the parameter $\gamma$ is empirically set to 0.2 in the experiments, and $d(i,j)$ represents the normalized Euclidean distance between the central point of $P_{l,i}$ and that of $P_{l,j}$. It can be seen that if two objects $P_{l,i}$ and $P_{l,j}$ are not adjacent, their value in the adjacency matrix is set to 0. Besides, diagonal elements are also set to 0. Through the operations, the structure information of neighbors is embedded into the feature information of nodes.

By combining the deep feature learning and object-based analysis techniques, the abovementioned procedure of graph construction not only takes the spatial and temporal information into account, but also utilizes the high level semantic features to represent the nodes. This strategy helps to efficiently and fully exploit kinds of information to achieve better performance. In addition, the interactions among adjacent nodes are modeled with their distances and similarities of the extracted features, which would be instrumental in accurately clustering the nodes for the consideration of structure information. The built features of nodes and the adjacency matrices represent a compact and effective way to provide relatively comprehensive information as input to the multiscale GCN.

### D. Multiscale GCN

Our GCN model is inspired from [53] and uses the proposed features of nodes and adjacency matrices as input. GCN extends the concept of convolution from regular grids data to graph structured data by generating node embeddings that gradually fuse the features in the neighborhood. The intrinsic difference grid-based convolution is that the number of neighbors of a node is not fixed in graph-based convolution. GCN can be categorized into two types: spectral and spatial based. In this paper, we employ a spectral-based approach in which the spectral filtering on graphs is defined.

The spectral convolutions on graphs can be defined as the multiplication of a signal $\mathbf{x} \in \mathbf{R}^N$ with a filter $g_\theta = diag(\theta)$ parameterized by $\theta \in \mathbf{R}^N$ in the Fourier domain, namely:

$$g_\theta * \mathbf{x} = \mathbf{U} g_\theta \mathbf{U}^T \mathbf{x} \quad (4)$$

where $\mathbf{U}$ is the matrix of eigenvectors of the normalized graph Laplacian $\mathbf{L} = \mathbf{I} - \mathbf{D}^{-1/2}\mathbf{A}\mathbf{D}^{-1/2} = \mathbf{U}\boldsymbol{\Lambda}\mathbf{U}^T$, with $\mathbf{A}$ being the adjacent matrix, $\boldsymbol{\Lambda}$ being the diagonal matrix containing the eigenvalues of $\mathbf{L}$, $\mathbf{I}$ being the identity matrix with proper size, $\mathbf{D}_{ii} = \sum_j \mathbf{A}_{ij}$. Then, the filter $g_\theta$ can be understood as a function of the eigenvalues of $\mathbf{L}$, i.e. $g_\theta(\boldsymbol{\Lambda})$. However, evaluating the formula (4) is computationally expensive. To reduce the computational cost of eigenvector decomposition, Hammond et al [65] approximated $g_\theta(\boldsymbol{\Lambda})$ using Chebyshev polynomials $T_k(\mathbf{x})$ up to $K$th order:

$$g_{\theta'}(\boldsymbol{\Lambda}) \approx \sum_{k=0}^{K} \theta'_k T_k(\tilde{\boldsymbol{\Lambda}}) \quad (5)$$

where $\boldsymbol{\theta}' \in \mathbf{R}^K$ is a vector of Chebyshev coefficients, and $\tilde{\boldsymbol{\Lambda}} = (2/\lambda_{max})\boldsymbol{\Lambda} - \mathbf{I}$ with $\lambda_{max}$ being the largest eigenvalue of $\mathbf{L}$, The Chebyshev polynomials are recursively defined as $T_k(\mathbf{x}) = 2\mathbf{x}T_{k-1}(\mathbf{x}) - T_{k-2}(\mathbf{x})$, with $T_0(\mathbf{x}) = 1$ and $T_1(\mathbf{x}) = \mathbf{x}$. Therefore, the convolution on graph signal $\mathbf{x}$ with a filter $g_{\theta'}$ can be defined as:

$$g_{\theta'} * \mathbf{x} \approx \sum_{k=0}^{K} \theta'_k T_k(\tilde{\mathbf{L}})\mathbf{x} \quad (6)$$

with $\tilde{\mathbf{L}} = 2/\lambda_{max}\mathbf{L} - \mathbf{I}$ denoting the scaled Laplacian matrix. Equation (6) can be easily verified by noticing that $(\mathbf{U}\boldsymbol{\Lambda}\mathbf{U}^T)^k = \mathbf{U}\boldsymbol{\Lambda}^k\mathbf{U}^T$. From (6), it can be inferred that the convolution on graph signal $\mathbf{x}$ depends only on nodes that are at maximum $K$ steps away from the central node ($K$th-order neighborhood). In this paper, we consider the first-order neighborhood, i.e., $K=1$, in another word, the central node and those adjacent to it are involved. Thus, (6) becomes a linear function on the graph Laplacian spectrum with respect to $\mathbf{L}$.

With the linear formulation, Kipf and Welling [53] further approximated $\lambda_{max} \approx 2$, as the neural network parameters can adapt to this change in scale during the training process.

We build a network of $L$ channels corresponding to the segmentations results and the $L$ constructed graphs, as the configuration shown in Fig.1. To be specific, equation (6) for the $l$th channel can be simplified to

$$g_{\theta'_l} * \mathbf{x} \approx \theta'_{l,0} + \theta'_{l,1}(\mathbf{L}_l - \mathbf{I}) = \theta'_{l,0} - \theta'_{l,1}\mathbf{D}_l^{-1/2}\mathbf{A}_l\mathbf{D}_l^{-1/2}\mathbf{x} \quad (7)$$

where the subscript $l$ denotes the $l$th channel ($l = 1, \cdots, L$), $\theta'_{l,0}$ and $\theta'_{l,1}$ are two free parameters. Since it can be beneficial to constrain the number of parameters to address overfitting and to minimize the number of operations (such as the matrix multiplications) per layer, (7) can be converted to

$$g_{\theta_l} * \mathbf{x} \approx \theta_l(\mathbf{I} + \mathbf{D}_l^{-1/2}\mathbf{A}_l\mathbf{D}_l^{-1/2})\mathbf{x} \quad (8)$$

by setting $\theta_l = \theta'_{l,0} = -\theta'_{l,1}$. Note that $\mathbf{I} + \mathbf{D}_l^{-1/2}\mathbf{A}_l\mathbf{D}_l^{-1/2}$ has eigenvalues in the range [0, 2], numerical instabilities and exploding/vanishing gradients may happen if this operator is repeatedly applied in a deep neural network. To alleviate this problem, we use the renormalization trick Kipf and Welling introduced in [53]: $\mathbf{I} + \mathbf{D}_l^{-1/2}\mathbf{A}_l\mathbf{D}_l^{-1/2} \to \tilde{\mathbf{D}}_l^{-1/2}\tilde{\mathbf{A}}_l\tilde{\mathbf{D}}_l^{-1/2}$, with $\tilde{\mathbf{A}}_l = \mathbf{A}_l + \mathbf{I}, \tilde{\mathbf{D}}_{l,ii} = \sum_j \tilde{\mathbf{A}}_{l,ij}$.

After that, we can generalize this definition to the signal $\mathbf{F}_l \in \mathbf{R}^{N_l \times C}$ introduced in Section III.C. Here, $N_l$ is the number of objects under the scale parameter $S_l$, and $C$ is the dimensional number of feature vector for each node which is 128 in this paper. The convolutional operation on the constructed graphs above therefore can be written as:

$$\mathbf{Z}_l = \tilde{\mathbf{D}}_l^{-1/2}\tilde{\mathbf{A}}_l\tilde{\mathbf{D}}_l^{-1/2}\mathbf{F}_l\mathbf{W}_l \quad (9)$$

where $\mathbf{Z}_l \in \mathbf{R}^{N_l \times M}$ is the convolved signal matrix, and $\mathbf{W}_l \in \mathbf{R}^{C \times M}$ is trainable weight matrix.

A multilayer model based on graph convolutions can therefore be built by stacking multiple convolutional layers of the form of (9) with the activation function. Considering $N$ layer GCN in the $l$th channel, the proposed forward model takes the following form:

$$f_l(\mathbf{F}_l, \mathbf{A}_l) = \text{softmax}\left(\tilde{\mathbf{A}}_l \cdots \left(\sigma\left(\tilde{\mathbf{A}}_l \sigma\left(\tilde{\mathbf{A}}_l \mathbf{F}_l \mathbf{W}_l^0\right)\right)\right) \cdots \mathbf{W}_l^{N-1}\right) \quad (10)$$

where $\mathbf{W}_l^0$ denotes the weight matrix from input to the first hidden layer and $\mathbf{W}_l^{N-1}$ is that from the last hidden layer to the output, $\sigma(\bullet)$ represents an activation function, such as the Relu function adopted in this paper. The softmax activation function, which is defined as $\text{softmax}(x_i) = \exp(x_i)/Z$ with $Z = \sum_i \exp(x_i)$.

### E. Fusion Module

According to the equation (10), the dimensional numbers of output $\mathbf{O}_1 = f_1(\mathbf{F}_1, \mathbf{A}_1), \cdots, \mathbf{O}_L = f_L(\mathbf{F}_L, \mathbf{A}_L)$ of the $L$ channels are $N_1 \times F, \ldots, N_L \times F$, respectively, with $F$ being the class number which is 2 in this paper. To preserve detailed information as much as possible in the final results, we design the fusing matrices for $\mathbf{O}_2, \cdots, \mathbf{O}_L$ to fuse them into the finest output $\mathbf{O}_1$.

As mentioned in Section III.A, one segmented object under $S_2, \cdots, S_L$ can be obtained by merging several objects under $S_1$ with other segmented parameters keeping invariant. We define the fusing matrices using the area ratio and spectral similarity between the child and father objects. The elements of fusing matrices $\mathbf{T}_l \in \mathbf{R}^{N_1 \times N_l}, l = 2, \cdots, L$ can be calculated as:

$$\mathbf{T}_{l,ij} = \frac{\#(P_{1,i})}{\#(P_{l,j})} \bullet \exp(-\beta \bullet sim(P_{1,i}, P_{l,j})) \quad (11)$$

where $\#(\bullet)$ represents the number of pixels in the object, the parameter $\beta$ is empirically set to 0.5 in the experiments, and $sim(P_{1,i}, P_{l,j})$ denotes the spectral similarity between $P_{1,i}$ and $P_{l,j}$ which is calculated using distance between the spectral mean vectors of the two objects.

After that, the coarser outputs $\mathbf{O}_2,\cdots,\mathbf{O}_L$ can be refined to the finest scale with the same size and object set as $\mathbf{O}_1$ using the multiplication of the fusing matrix with $\mathbf{O}_1$. Thus, the fusion of the multi-channel output results can be obtained with the following formula:

$$\mathbf{E} = \mathbf{O}_1 + \mathbf{T}_2\mathbf{O}_2 + \cdots + \mathbf{T}_L\mathbf{O}_L \tag{12}$$

where the fused result $\mathbf{E}$ has the same size and corresponding nodes with $\mathbf{O}_1$.

The aforementioned fusion module is designed based on the following considerations. First, the finest result can preserve more details, whereas, reasonably adding coarser information would reduce the error detection for the reason that only the finest one can hardly present the ground objects with various sizes. The definition of fusing matrix is consistent with the intuition of vision for the element in the matrix is large when the spectral values of the child and father objects are close, and the area of the child accounts for a great proportion in that of the father. Second, only a fraction of the objects under the finest scale need to be labeled (as the red nodes shown in Fig.1), which may evidently reduce the workloads and cost of labeling samples.

We then evaluate the cross-entropy error as the loss function:

$$\mathcal{L} = -\sum_{t \in Y_T} \sum_{f=1}^{F} Y_{tf} \ln \mathbf{E}_{tf} \tag{13}$$

where $Y_T$ is the set of node indices that have labels. Given $\mathcal{L}$, the network weights $\mathbf{W}_l^0,\cdots,\mathbf{W}_l^{N-1}$ $(l=1,\cdots,L)$ are trained using gradient descent approach, where all the nodes are utilized to perform gradient descent. In this way, the gradient information from the labeled nodes is spread across other unlabeled adjacent nodes.

## IV. EXPERIMENTS AND DISCUSSION

In this section, the illustration of data sets is firstly presented. Then, we provide a brief description of the implementation details and evaluation metrics. Following that, the experimental results performed on the optical, SAR and optical-SAR data sets are presented and analyzed, respectively. Finally, some discussion about our method is made in detailed.

### A. Descriptions of Data Sets

For the optical cases, two openly VHR RS image data sets are employed: the Beijing & Tianjin data set provided by B. Hou and Q. Liu [36] and the LEVIR-CD data set [66] released by H. Chen and Z. Shi. The first data set includes 29 pairs of images covering two big cities, Beijing and Tianjin, in China during the period from 2006 to 2017. These images are captured at different times of a day and seasons under different imaging conditions, which increase the complexities and diversities of the data. They are with quite large sizes, e.g., 2000×2000 pixels, most of these images are collected by Google Earth with the spatial resolution of 0.46m, some test image patches collected by GF-2 with the spatial resolution of 1m are also provided. Several patch samples of the first data set are shown in Fig.3. The second data set consists of 637 pairs of

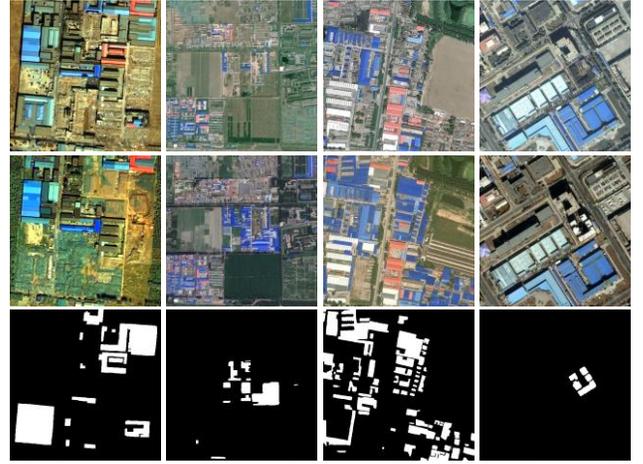

Fig.3. Image patch examples and corresponding reference images of the Beijing &Tianjin data set.

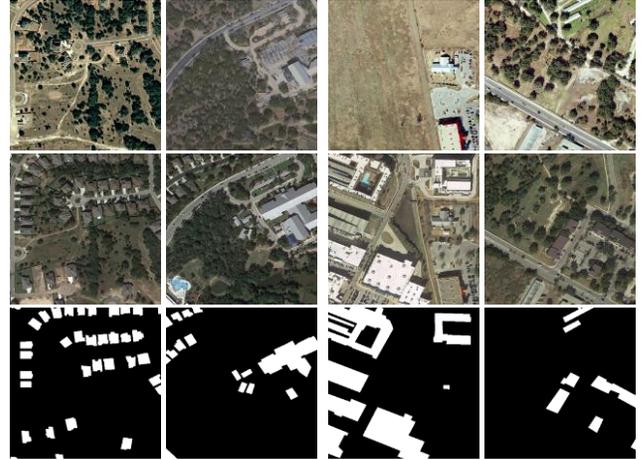

Fig.4. Image patch examples and corresponding reference images of the LEVIR-CD data set.

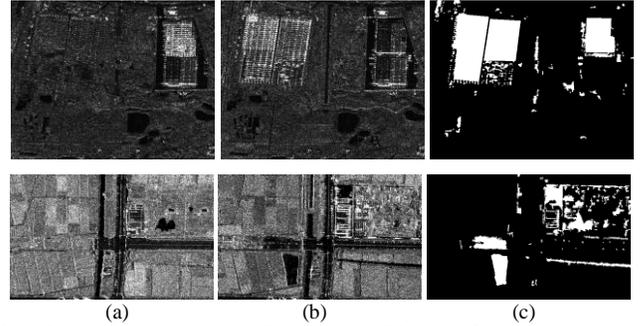

(a)  (b)  (c)
Fig.5. SAR data sets. (a) Image T1. (b) Image T2. (c) Reference change map

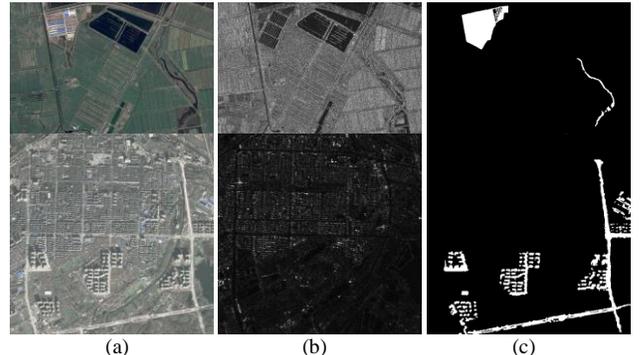

(a)  (b)  (c)
Fig.6. Heterogeneous data sets. (a) Optical image. (b) SAR image. (c) Reference change map

images with sizes of 1024×1024 and the spatial resolution of 0.5m collected by Google Earth, covering different cities in Texas of the US, including Austin, Lakeway, Bee Cave, Buda, Kyle, Manor, Pflugervilletx, Dripping Springs, etc. Some examples of the second data set are shown in Fig.4. The two data sets mainly represent the construction and decline of buildings. Meanwhile, both of them introduce variations derived from the seasonal factors and illumination conditions, which could help develop effective methods that can mitigate the impact of irrelevant changes on real changes.

Two VHR data sets are used to conduct the experiments of CD for SAR images. The first data set is a pair of images acquired by TerraSAR-X sensor with HH polarization and 1m/pixel covering a suburban area of Wuhan, China, where the remarkable changes are the construction and demolition of buildings, as shown in the first row Fig.5. The second data set corresponds to an area in Beijing, China, as shown in the second row in Fig5. The images are acquired by Gaofen-3 with the size of 550×900 and a spatial resolution of 1m/pixel.

Two heterogeneous data sets are used to evaluate the efficiency of the proposed method in heterogeneous CD tasks. The first data set is a pair of optical/SAR images (Shuguang Village, Dongying, China), as shown in the first row of Fig.6. The optical image, with the size of 593×921 and red, green, blue bands, is acquired from Google Earth in September 2012; whereas the SAR image is taken by the Radarsat-2 with the size of 593×921 and C-band in June 2008. The changes of land use in the farmland and water cover are the most significantly changed information during the period. The second pair of optical/SAR images shown in the second row of Fig. 6 describe an urban area in Wuhan, China. The optical image was captured from Google Earth with the size of 495×503 and red, green, blue bands in November 2011; whereas the SAR image is acquired by Radarsat-2 with the size of 495×503 and C-band in June 2008.

### B. Implementation and Evaluation Metrics

The proposed method is implemented via Pytorch framework on a single GeForce GTX 1080Ti GPU. In the feature extracting network training phase, images with labels in the OSCD dataset are clipped into 6400 training images of 112×112 pixels with data augmentation, including rotation, flip. The stochastic gradient descent (SGD) with momentum is applied for training. The learning rate is fixed and set to 0.001, meanwhile, the momentum and the weight decay is set to 0.9 and 0.0005, respectively. For each image pair introduced in Section IV-A, we randomly select 5% of the objects under the finest scale as labeled nodes. The number of GCN layers is set to 3, namely $N=3$ in formula (10). We employ 3 scales to conduct the FNEA, namely $L=3$ in formula (12) and Fig.1. We train the proposed MSGCN for 400 epochs with the dropout rate of 0.5 and the weight decay of 0.0005.

To evaluate the performance of the proposed method, four quantitative evaluation indices, false alarm rate (FAR), missed alarm rate (MAR), overall accuracy (OA) and Kappa coefficient (Kappa) are adopted as metrics. The equations of FAR, MAR and OA can be formulated as FAR=FP/(FP+TN), MAR=FN/(FN+TP) and OA=(TP+TN)/(TP+TN+FP+FN), respectively, where TP denotes the number of true positives, FP denotes the number of false positives, TN denotes the number of true negatives, and FN denotes the number of false negatives. Kappa is a statistical measure of the consistency between the change map and the reference map. It is calculated by

$$\text{Kappa}=(OA-PRE)/(1-PRE),$$
$$PRE=\frac{(TP+FN)g(TP+FP)+(TN+FP)g(TN+FN)}{(TP+TN+TP+FN)^2} \quad (14)$$

### C. Experiments on Optical Images

To verify the effectiveness of the proposed method for optical images, we compare our method with the following six benchmark methods on the aforementioned Beijing & Tianjin and LEVIR-CD data sets:

1) FC-Siam-con: The fully convolutional Siamese concatenation (FC-Siam-con) [61] applied a Siamese encoding stream to extract deep features from bi-temporal images, then, the features were concatenated in the decoding stream for CD.

2) FCN-PP: The fully convolutional network with pyramid pooling (FCN-PP) [67] has been proposed for landslide detection. It consists of a U-shape architecture to learn the deep features of the input images and a pyramid pooling layer to enlarge the receptive filed.

3) Unet_ASPP: Unet is a widely used architecture in the semantic segmentation and CD tasks. The atrous spatial pyramid pooling (ASPP) can capture contextual information with multi-rate via sampling the input feature map using dilated convolutions of different sampling rates in parallel [68]. We insert the ASPP module between the down-sampling and up-sampling operators in Unet, that is, Unet_ASPP.

4) DSIFN: A deeply supervised image fusion network (DSIFN) has been proposed for CD in high resolution RS images [69]. The DSIFN consists of a shared deep feature network and a difference discrimination network which utilizes the channel attention module and spatial attention module. The authors have release the codes in [69].

5) W-net: The W-shape network proposed in [36] applied an end-to-end dual-branch architecture, and performed the differencing operator in the feature domain rather than in the traditional image domain, which greatly alleviated loss of useful information for determining the changes.

6) GCNCD: A network with two GCN layers has been proposed for CD in [58]. The method uses several hand-crafted features of objects to build nodes of the graph model.

The first five methods are supervised, and we use 1200 pairs of images for training, 40 pairs for testing on the Beijing & Tianjin data set and 1780 pairs for training, 512 pairs for testing on the LEVIR-CD data set, respectively, to evaluate these methods. The sizes of all images above are 512×512. The GCNCD is semi-supervised as our method. Thus, to ensure the fairness, we randomly select 5% of the objects under the finest scale as labeled nodes for the GCNCD and our method. The three scale parameters of multiscale segmentation are set as 8, 15, and 20 for both Beijing & Tianjin and LEVIR-CD data sets.

Some of the results on the Beijing & Tianjin testing data are presented in Fig.7. Similarly, some typical results on the LEVIR-CD testing data are displayed in Fig.8.

Intuitively, the change maps generated by the proposed are more consistent with the reference change maps. To be specific, there are many discontinuous small noise strips (false alarms) in the results of FC-Siam-con, caused by its limited robustness

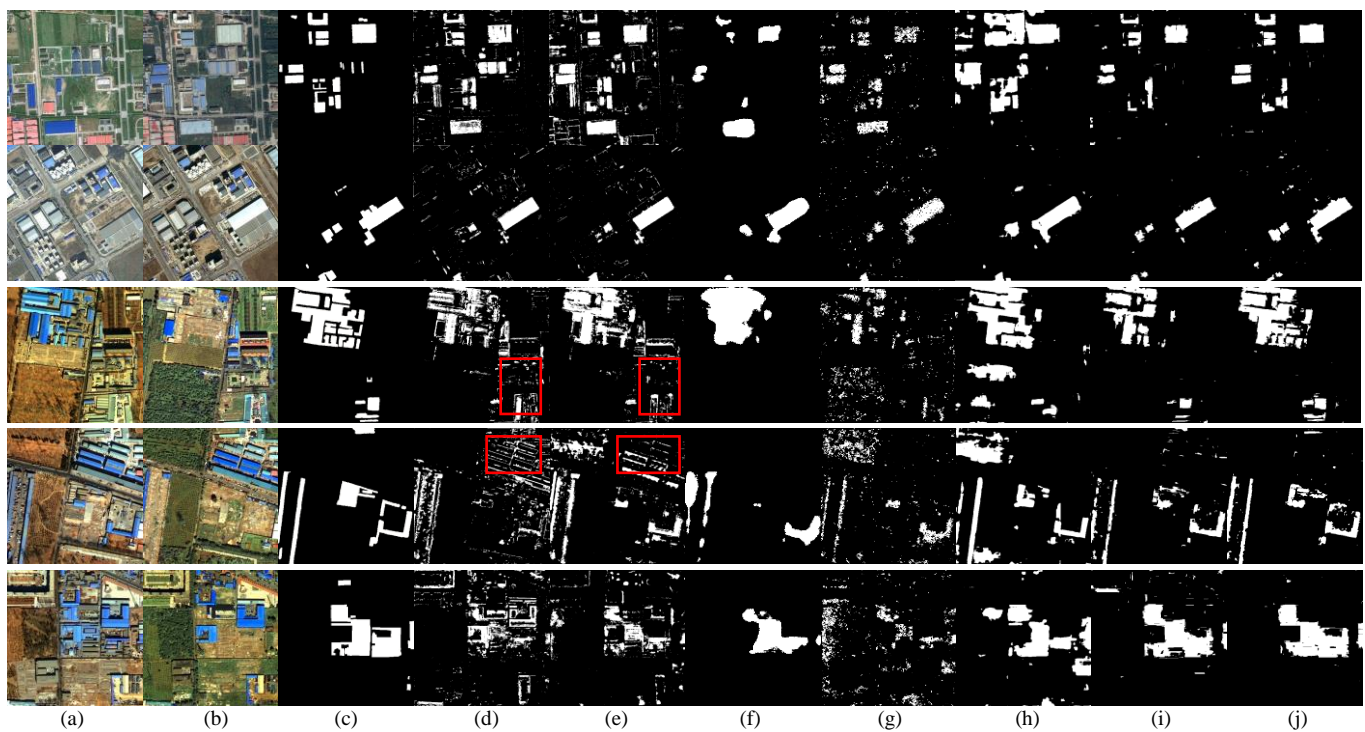

(a) (b) (c) (d) (e) (f) (g) (h) (i) (j)

Fig.7. Some typical CD maps by different methods on the Beijing & Tianjin data set. (a) Image T1. (b) Images T2. (c) Reference change map. (d) FC-Siam-con. (e) FCN-PP. (f) Unet_ASPP. (g) DSIFN. (h) W-net. (i) GCNCD. (j) Proposed MSGCN.

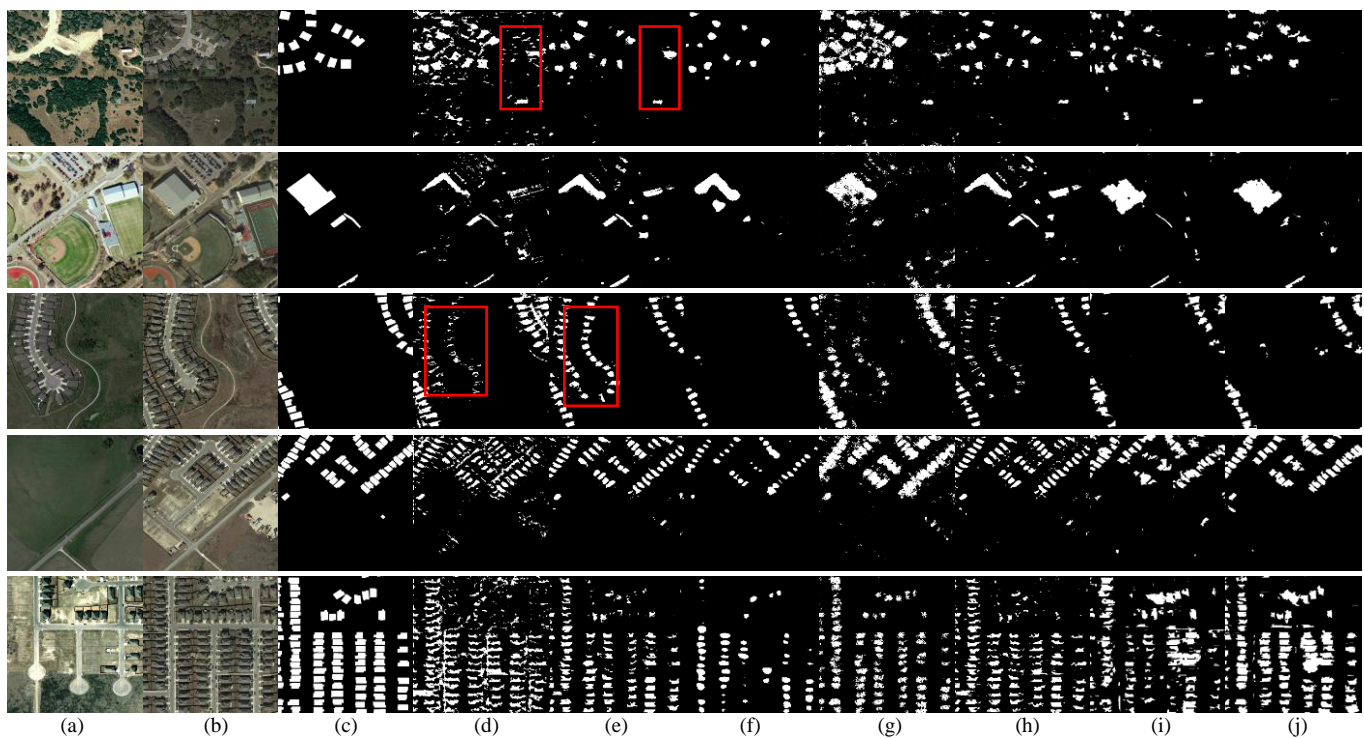

(a) (b) (c) (d) (e) (f) (g) (h) (i) (j)

Fig.8. Some typical CD maps by different methods on the LEVIR-CD data set. (a) Image T1. (b) Images T2. (c) Reference change map. (d) FC-Siam-con. (e) FCN-PP. (f) Unet_ASPP. (g) DSIFN. (h) W-net. (i) GCNCD. (j) Proposed MSGCN.

on the inevitable misregistration errors and spectral variations in high resolution RS image pairs. Due to the similar reasons, FCN-PP causes massive false alarms, as can be seen in the red boxes in Fig.7(d), Fig.7(e), Fig.8(d) and Fig.8(e). The Unet_ASPP can cause fewer false alarms and obtain smoother results. However, the excessive smoothness hinders its performance at areas with complex structures. Thus, the accuracies of boundaries between changed and unchanged regions are dissatisfactory. Besides, as is clear, some small changed regions are missed in the results of Unet_ASPP. The results of DSIFN exit considerable false alarms and missed detections. The W-net achieves relatively good performances on some image pairs, such as the second row of Fig.7(h) and the fourth row of Fig.8(h), but obvious false alarms occur in some results ,such as the fourth row of Fig.7(h) and the third row of Fig.8(h). The results of GCN have good homogeneity in changed regions and obtain relatively precise boundaries. Nevertheless, due to the fact that hand-crafted features have limited ability of presentation, GCN fails to capture some complex structures of changed regions, such as the third row of Fig.7(i). The proposed MSGCN can reduce the false alarms and missed detection to a low level, simultaneously, by incorporating the deep features with high robustness and the multiscale object analysis into GCN.

TABLE I
QUANTITATIVE ACCURACY RESULTS FOR DIFFERENT METHODS ON THE OPTICAL DATA SETS (%)

| Data set | Method | FAR | MAR | OA | Kappa |
|---|---|---|---|---|---|
| Beijing &Tianjin | FC-Siam-con | 7.08 | 59.54 | 86.52 | 47.39 |
| | FCN-PP | 6.19 | 45.89 | 89.13 | 52.81 |
| | Unet_ASPP | **0.90** | 52.04 | 88.02 | 51.17 |
| | DSIFN | 4.36 | 46.15 | 90.23 | 56.90 |
| | W-net | 3.45 | 38.01 | 91.01 | 59.58 |
| | GCNCD | 1.69 | 26.35 | 94.92 | 71.10 |
| | MSGCN | 1.42 | **21.32** | **96.71** | **77.58** |
| LEVIR-CD | FC-Siam-con | 6.23 | 41.91 | 88.57 | 56.36 |
| | FCN-PP | 3.94 | 32.12 | 90.61 | 63.58 |
| | Unet_ASPP | **0.31** | 42.45 | 89.98 | 61.32 |
| | DSIFN | 2.42 | 25.21 | 93.89 | 67.52 |
| | W-net | 3.56 | 30.01 | 91.12 | 64.35 |
| | GCNCD | 2.31 | 26.79 | 92.80 | 66.88 |
| | MSGCN | 1.13 | **23.89** | **95.68** | **74.79** |

The quantitative evaluation results of different methods are listed in Table I. It can be concluded that the quantitative results are consistent with the visual performance, and the proposed MSGCN outperforms all of the compared methods in terms of Kappa, OA and MAR on both of the two data sets. Regarding the FAR, the MSGCN is the second best. Comparing with other methods, the MSGCN yields an improvement of at least 5.02%, 1.19%, and 6.48% for MAR, OA and Kappa, respectively, on the Beijing & Tianjin data set. The improvement on the LEVIR-CD data set is also evident, with the improving values of at least 2.90%, 1.79%, and 7.27% for MAR, OA and Kappa, respectively. The Unet_ASPP achieves slightly lower FARs than those of MSGCN. However, it yields significantly higher MARs. On the whole, the MSGCN can suppress the false alarms and reduce the missed detections, simultaneously. The reasons of this behavior are: 1) the combination of pixel-wise high level features with object-based extraction improve the robustness on the misregistration errors and spectral variations, and this can reduce the false alarms and missed detection; 2) the multiscale features are made full use in MSGCN, thus, information from the labeled nodes can be spread to the unlabeled ones, effectively and accurately.

*D. Experiments on SAR Images*

We evaluate the performance of MSGCN on two SAR data sets comparing with the following four state-of-the-art CD methods:

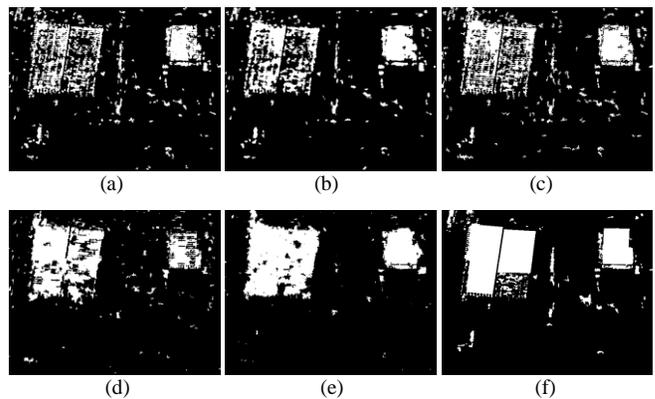

Fig.9. Visual results by different methods on the Wuhan SAR data set. (a) PCA-Net. (b) S-PCA-Net. (c) CWNN. (d) CNN. (e) Proposed MSGCN. (f) Reference change map.

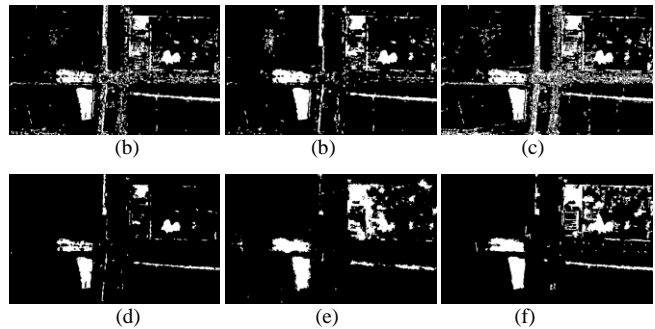

Fig.10. Visual results by different methods on the Beijing SAR data set. (a) PCA-Net. (b) S-PCA-Net. (c) CWNN. (d) CNN. (e) Proposed MSGCN. (f) Reference change map.

1) PCA-Net. The PCA-Net has been applied to SAR image CD [70].The main difference from convolution neural network is that the filter kernels are obtained without back-propagation, instead, they are considered as the eigenvectors of most large eigenvalues after applying the eigen decomposition to covariance matrix.

2) S-PCA-Net. For SAR image CD, the S-PCA-Net introduced the imbalanced learning process into PCA-Net to solve the imbalance samples issue [71].

3) CWNN. The method used convolutional-wavelet neural network (CWNN) instead of CNN to extract robust features with better noise immunity for SAR image CD [72].

4) CNN. The method proposed a novel CNN framework without any preprocessing operations, which can automatically extract the spatial characteristics [73], and the codes have been released at https://github.com/xhwNobody/Change-Detection.

Among these methods, PCA-Net and CWNN are unsupervised methods, while S-PCA-Net and CNN are supervised ones. The involved parameters of these methods are set as those in their original articles. From fine to coarse, the three scale parameters of multiscale segmentation are set as 10, 15, 25 for Wuhan SAR data set and 15, 25, 35 for Beijing SAR data set, respectively.

The visual results on the two SAR data sets are shown in Fig.9 and Fig.10. It is shown that for the Wuhan data set, the proposed MSGCN can obtain more completed changed regions. Particularly, much more changed regions are missed in the left building construction area in Fig.9(a)~ Fig.9(d) comparing with the result of MSGCN. In addition, many isolated spots appear in Fig.9(a)~ Fig.9(d). By contrast, the MSGCN can reduce the number of isolated spots, as shown in Fig9(e). For Beijing data set, due to the influence of speckle noise, the PCA-Net, S-PCA-Net and CWNN cause many false alarms. To be specific, the horizontal road haven't been changed during the two imaging times, but many pixels of the road are detected as changes in Fig.10(a)~ Fig.10(c). In comparison, both CNN and MSGCN can effectively suppress false alarms. However, quite a lot of changed regions are missed in Fig.10(d), such as the up-right part, where building constructions had happened during the two imaging times, but the CNN method fails to accurately detect this changed information. On the whole, the MSGCN can produce change map with completed changed regions and relatively less false alarms.

TABLE II
QUANTITATIVE ACCURACY RESULTS FOR DIFFERENT METHODS ON THE SAR DATA SETS (%)

| Data set | Method | FAR | MAR | OA | Kappa |
|---|---|---|---|---|---|
| Wuhan | PCA-Net | 3.34 | 35.84 | 91.71 | 66.12 |
|  | S-PCA-Net | **2.71** | 32.91 | 92.54 | 70.58 |
|  | CWNN | 3.61 | 28.92 | 92.41 | 71.21 |
|  | CNN | 3.41 | 28.87 | 92.59 | 71.78 |
|  | MSGCN | 3.05 | **13.09** | **95.37** | **82.77** |
| Beijing | PCA-Net | 6.44 | 40.78 | 90.17 | 48.87 |
|  | S-PCA-Net | 4.43 | 39.74 | 92.08 | 55.65 |
|  | CWNN | 10.97 | 36.50 | 86.51 | 40.93 |
|  | CNN | **0.76** | 57.08 | 93.68 | 54.27 |
|  | MSGCN | 2.70 | **31.02** | **94.50** | **62.11** |

Table II shows the quantitative evaluation results on the two SAR data sets. It can be seen that the MSGCN outperform other methods significantly, with the improving values of at least 15.78%, 2.78%, and 10.99% for MAR, OA and Kappa on the Wuhan data set, and at least 5.48%, 0.82%, and 6.46% for MAR, OA and Kappa on the Beijng data set, respectively. For the Wuhan data set, although the MSGCN achieves the second lowest FAR, which is a little higher than the S-PCA-Net, the MAR is reduced from 32.91% to 13.09% by a large margin. For the Beijing data set, due to the complicated noise situation, the performances of the unsupervised PCA-Net and CWNN are not very well. Obviously, the false alarms in results of these two methods are much more than others. Although the CNN achieves slightly lower FAR than the MSGCN, it yields significantly higher MARs, which is as high as 57.08%, and this means that many changed pixels are missed in the result of CNN. Comparing the indices in Table II comprehensively, the MSCGN produces best results on both data sets.

*E. Experiments on Heterogeneous Optical/SAR Images*

To evaluate the efficiency of the proposed MSGCN on heterogeneous optical/SAR image CD tasks, we validate it comparing with the following benchmark methods:

1) HPT. The homogeneous pixel transformation (HPT) [29] estimated mapping pixels based on the known unchanged pixels. In our experiments, we use 40% of the unchanged pixels as training samples.
2) M3CD. The Markov model for multimodal change detection (M3CD) algorithm used an observation field built up from a pixel pairwise modeling in an unsupervised way [74].

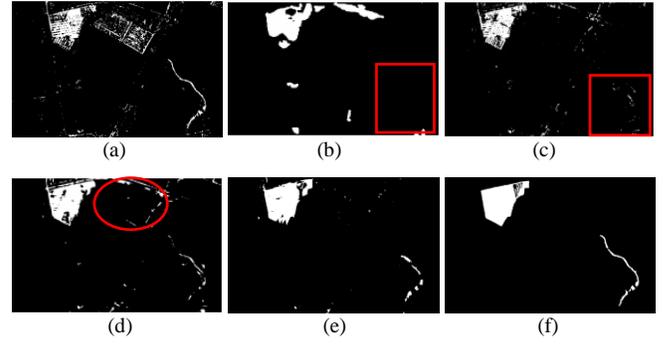

Fig.11. Visual results by different methods on the Shuguang heterogeneous data set. (a) HPT. (b) M3CD. (c) SCCN. (d) PSGM. (e) Proposed MSGCN. (f) Reference change map.

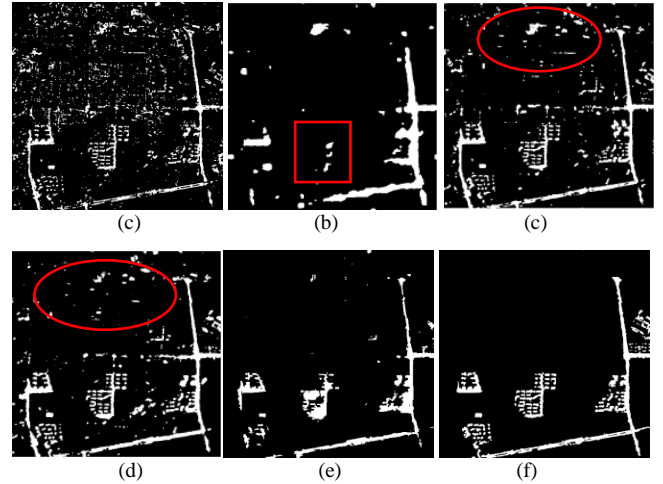

Fig.12. Visual results by different methods on the Wuhan heterogeneous data set. (a) HPT. (b) M3CD. (c) SCCN. (d) PSGM. (e) Proposed MSGCN. (f) Reference change map.

3) SCCN. A symmetric convolutional coupling network (SCCN) [75] was designed to infer spatial information form the data and learn new representations for heterogeneous CD.
4) PSGM. The unsupervised patch similarity graph matrix-based (PSGM) [76] method assumed that the patch similarity graph structure of each homogeneous or heterogeneous image is consistent if no change occurs.

From fine to coarse, the three scale parameters of multiscale segmentation are set as 10, 15, 20 for Shuguang data set and 7, 12, 20 for Wuhan data set, respectively.

Fig.11 and Fig.12 show the changed maps of all comparing methods on the heterogeneous Shuguang and Wuhan data set, respectively.

As can be seen, many unchanged pixels are misclassified into changes in Fig.11(a) and Fig.12(a), as the discrepancy between optical and SAR feature spaces can't be eliminated completely through the HPT. Therefore, the features of some unchanged ground objects are still dissimilar in the mapping feature space. The results of M3CD are much smoother than others, intuitively. However, excessive smoothing of the Markov model may reduce the accuracy of boundaries between changed and unchanged regions. Besides, some evidently changed areas are missed in the results of M3CD, such as the areas marked by the red boxes in Fig.11(b) and Fig.12(b), where the changes of water cover and constructions of some buildings are missed, respectively. The SCCN also misses many changed pixels in the area marked by the red box in Fig.11(c). In addition, relatively many false alarms are generated on the Wuhan data set, as shown in the red ellipse in Fig.12(c). The PSGM can obtain relatively complete changed information with some false alarm areas, such as the areas marked by red ellipses in Fig.11(d) and Fig.12(d). In contrast, the MSGCN generates obviously less false alarms on both data sets. Meanwhile, the completeness of changed regions can be well preserved by the MSGCN.

TABLE III
QUANTITATIVE ACCURACY RESULTS FOR DIFFERENT METHODS ON THE HETEROGENEOUS OPTICAL/SAR DATA SETS(%)

| Data set | Method | FAR | MAR | OA | Kappa |
|---|---|---|---|---|---|
| Shuguang | HPT | 2.28 | 35.91 | 96.17 | 58.62 |
|  | M3CD | 2.39 | 28.47 | 96.20 | 60.17 |
|  | SCCN | 0.93 | 36.21 | 97.18 | 65.20 |
|  | PSGM | 1.50 | 22.21 | 97.66 | 74.38 |
|  | MSGCN | **0.54** | **12.21** | **98.92** | **87.63** |
| Wuhan | HPT | 6.03 | 37.27 | 91.94 | 46.15 |
|  | M3CD | 3.55 | 40.20 | 93.38 | 47.58 |
|  | SCCN | 4.94 | 27.88 | 94.14 | 61.08 |
|  | PSGM | 4.12 | **24.01** | 95.32 | 68.58 |
|  | MSGCN | **1.75** | 30.40 | **96.38** | **69.58** |

The quantitative evaluation results on the two heterogeneous data sets are listed in Table III. For the Shuguang data set, the MSGCN outperforms other methods significantly in terms of all evaluation metrics. As it is consistent with the visual comparison of Fig.11, FAR and MAR of the MSGCN are both lower than those of others, and it means that the MSGCN can effectively suppress the false alarms and avoid the missed detections simultaneously. For the Wuhan data set, the MSGCN produces the best FAR, OA, and Kappa coefficient. Although SCCN and PSGM achieve lower MAR, they yield significantly higher FARs. In accordance with the visual analysis of Fig.12, many unchanged pixels are erroneously detected as changes in the results of SCCN and PSGM. On the whole, MSGCN still outperforms SCCN and PSGM, and the highest OA and Kappa coefficient of MSGCN can support this conclusion.

*F. Discussion*

In the following, effectiveness of the fusion module, Influence of the ratio of labeled samples, performance with different number of graph convolutional layers will be discussed in detail.

*1) Effectiveness of the fusion module*: In order to capture more comprehensive information of the ground objects with various sizes, we segment the input image pair under three scale parameters, which can be named as fine, medium, and coarse scale, respectively. A fusion module is designed to make full use of outputs of the multiscale GCN. As the finest segments can preserve more details and only a fraction of the objects under the finest scale need to be labeled, performances with four scale combinations (fine, fine-medium, fine-coarse and fine-medium-coarse) are compared to prove the effectiveness of the fusion module. Specifically, the fine-medium and fine-coarse combinations mean $L=2$ in formula (12), while the fine and fine-medium-coarse mean $L=1$ and $L=3$, respectively.

Take the Beijing SAR data set for instance, the three scale parameters are 15, 25 and 35, thus, the fine-medium case means the combination of 15 and 25.

Fig.13 shows two samples of optical CD results with different scale combinations, while Fig.14 and Fig.15 show the experimental results on the Wuhan SAR and heterogeneous data sets, respectively. The corresponding quantitative evaluation results are listed in Table IV. From Fig.13~15, interpreting in detail, we have the following observations. First, the fusion module is capable of capturing more accurate structures of changed regions. For instance, comparing the areas of red boxes in Fig.13, the changed regions of fine-medium-coarse combination is closer to the reference than those of single scale or two scale combinations. Second, incorporating multiscale information can alleviate the missed detection phenomenon, such as the areas of red boxes in Fig.14, where the holes in the detected changed regions are missed detections and the fine-medium-coarse combination obtains less and smaller holes than those of others. Since the ground objects in images can be of various sizes, considering a multiscale fusion strategy can be highly beneficial.

It can be observed from Table IV that fine-medium-coarse combination achieves the best performances in terms of MAR, OA and Kappa. In addition, fine-medium and fine-coarse combinations outperform the single fine scale on all data sets. The reason may be that the information that should be multiscale and it cannot be sufficiently exploited using only a single scale, especially when image scenes are complex. These demonstrate that the proposed fusion module can better exploit the multiscale information to improve the performances.

*2) Influence of the ratio of labeled samples*: Unlike other supervised methods which use some individual pairs of images as training data and other pairs as testing data, the semisupervised MSGCN performs training with a few labeled superpixels (objects) and rest unlabeled ones on each pair. Therefore, the ratio of labeled samples would unavoidably influence the performance of MSGCN and this influence should be investigated. To this end, we vary the ratio of labeled samples from 5% to 30% in steps of 5% and report the OA and Kappa coefficient on all the aforementioned data sets, as shown in Fig.16. We can make the observation from Fig.16 that the performance on all data sets can be improved by increasing the ratio of labeled examples. It is noteworthy that the proposed MSGCN can yield relatively high precision accuracy even though the labeled ratio is low, as the experiments in Section IV-C~ Section IV-E, where the MSGCN achieves promising performance with the labeled ratio of 5%. This advantage reduces the requirement of abundant labeled samples, and thus, makes it quite feasible in practical CD tasks.

*3) Performances with different numbers of graph convolutional layers*: The number of graph convolutional layers is an important hyperparameter in network design. To evaluate the performances with different numbers of layers, we implement the proposed MSGCN with two, three, four, and five graph convolutional layers, respectively. As the input dimension is 128 and the output dimension is 2, constantly, the dimensions of the two-layer, three-layer, four-layer and five-layer MSGCNs are set as 128-32-2, 128-32-8-2, 128-32-16-4-2 and 128-32-16-8-4-2, respectively. Fig.17 shows two samples of optical CD results with different layer

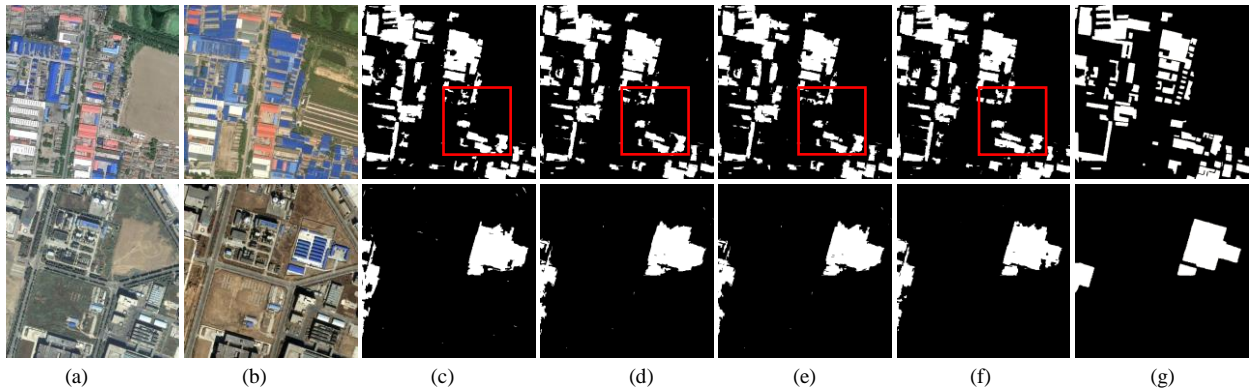

Fig.13. Two samples of optical CD results with different scale combinations. (a) Image T1. (b) Image T2. (c) Scale combination: 8. (d) Scale combination: 8-15. (e) Scale combination: 8-20. (f) Scale combination: 8-15-20. (g) Reference change map

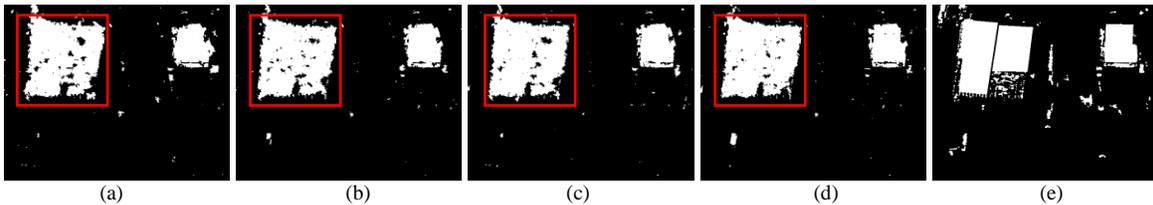

Fig.14. Change maps of Wuhan SAR data set with different scale combinations. (a) Scale combination: 10. (b) Scale combination: 10-15. (c) Scale combination: 10-25. (d) Scale combination: 10-15-25. (e) Reference change map

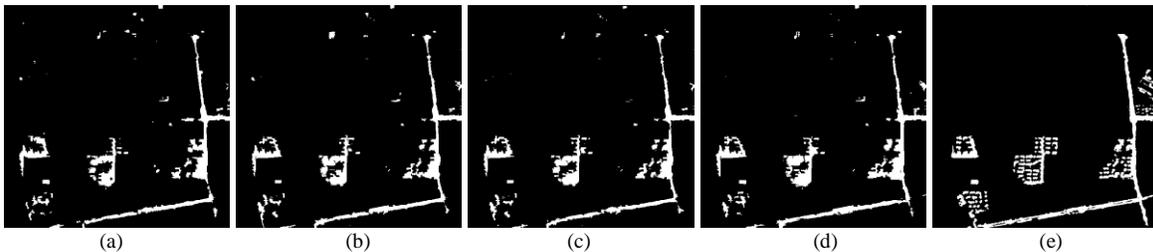

Fig.15. Change maps of Wuhan heterogeneous data set with different scale combinations. (a) Scale combination: 7. (b) Scale combination: 7-12. (c) Scale combination: 7-20. (d) Scale combination: 7-12-20. (e) Reference change map

TABLE IV
QUANTITATIVE ACCURACY RESULTS FOR DIFFERENT SCALE COMBINATIONS (%)

| Scale combinations | Optical images | | | | Wuhan SAR data set | | | | Wuhan heterogeneous data set | | | |
|---|---|---|---|---|---|---|---|---|---|---|---|---|
| | FAR | MAR | OA | Kappa | FAR | MAR | OA | Kappa | FAR | MAR | OA | Kappa |
| fine | 4.24 | 15.52 | 94.22 | 76.44 | 3.27 | 14.83 | 94.91 | 81.00 | 1.60 | 37.32 | 96.07 | 65.45 |
| fine-medium | 3.61 | 15.04 | 94.79 | 78.84 | **2.84** | 14.97 | 95.25 | 82.11 | 1.65 | 35.26 | 96.16 | 66.67 |
| fine-coarse | 3.58 | 16.09 | 94.77 | 78.77 | 3.25 | 14.11 | 95.04 | 81.55 | **1.45** | 37.30 | 96.22 | 66.37 |
| Fine-medium-coarse | **3.41** | **13.33** | **95.23** | **80.75** | 3.05 | **13.09** | **95.37** | **82.77** | 1.75 | **30.40** | **96.38** | **69.58** |

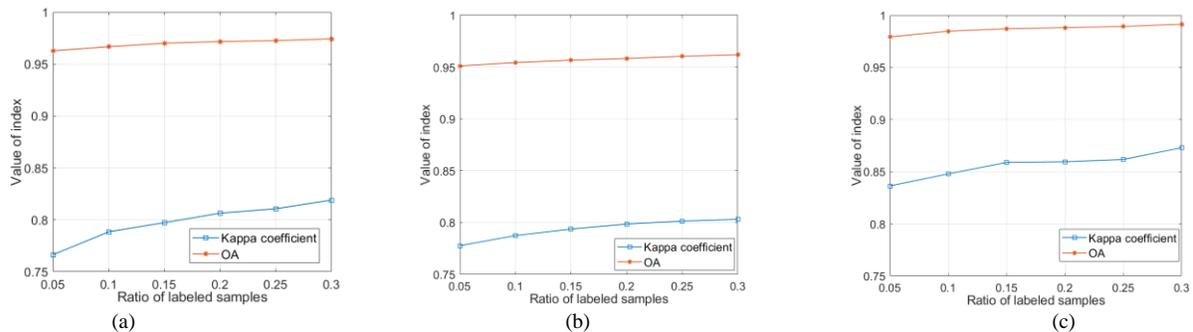

Fig.16. The Kappa coefficient and OA of MSGCN with different labeled ratios. (a) Optical images data sets. (b) SAR data set. (c) Heterogeneous data sets

numbers, while Fig.18 and Fig.19 show the experimental results on the Wuhan SAR and Shuguang heterogeneous data sets, respectively. Table V presents the corresponding quantitative evaluation results. The labeled ratios in these experiments are all set as 5%.

It can be observed from Fig.17~Fig.19 that increasing the layer number don't consequentially result in improvement of performance. For example, in the results of Shuguang heterogeneous data set, the obvious changes of water cover which locate in the right-down part are completely missed in the result with four layers, whereas, most of these changes can be captured in the result with two layers and three layers, as the regions marked by red boxes in Fig.19 (a)~(c). Interpreting in detail, results of three-layer MSGCN seen to be closer to the reference change maps on all the above data sets.

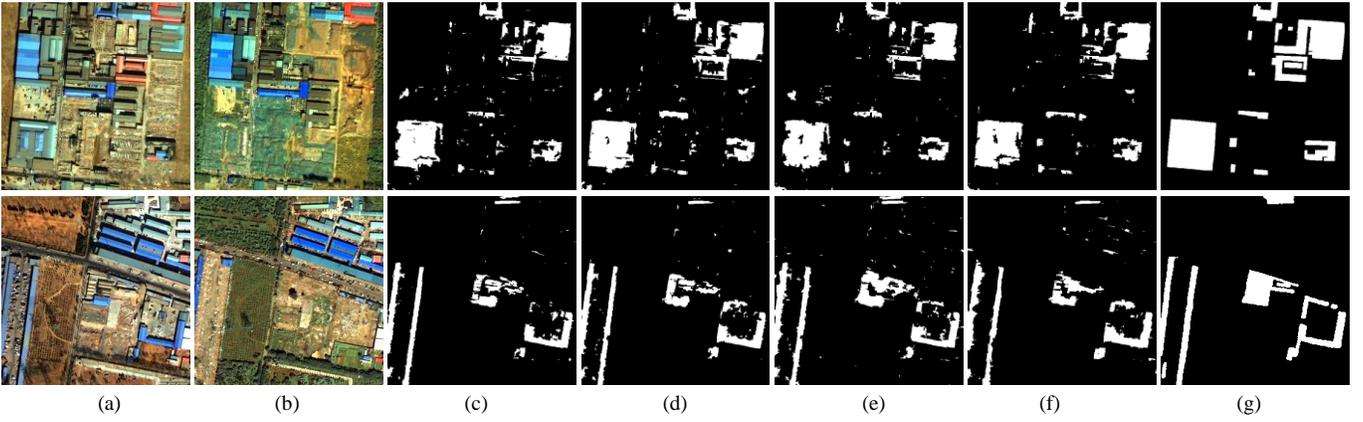

Fig.17. Two samples of optical CD results with different layer numbers. (a) Image T1. (b) Image T2. (c) Layer numbers: 2. (d) Layer numbers: 3. (e) Layer numbers: 4. (f) Layer numbers: 5. (g) Reference change map

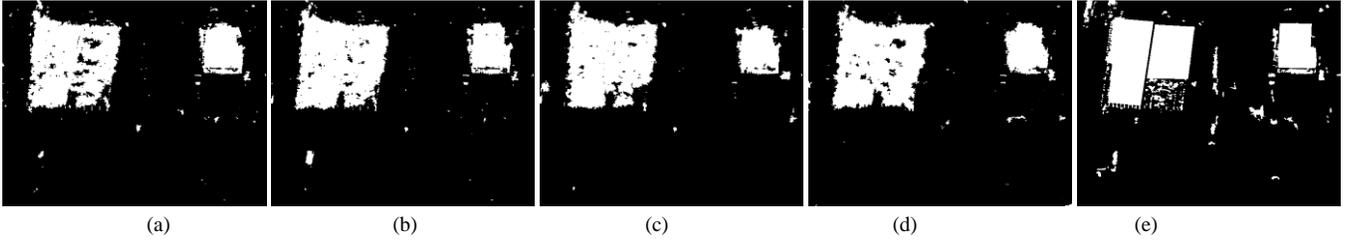

Fig.18. Change maps of Wuhan SAR data set with different layer numbers. (a) Layer numbers: 2. (b) Layer numbers: 3. (c) Layer numbers: 4. (d) Layer numbers: 5. (e) Reference change map

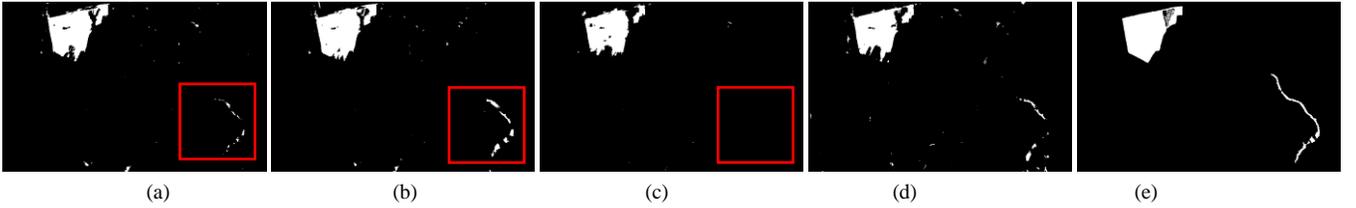

Fig.19. Change maps of Wuhan heterogeneous data set with different layer numbers. (a) Layer numbers: 2. (b) Layer numbers: 3. (c) Layer numbers: 4. (d) Layer numbers: 5. (e) Reference change map

TABLE V
QUANTITATIVE ACCURACY RESULTS FOR DIFFERENT NUMBERS OF GCN LAYERS (%)

| Number of layers | Optical images | | | | Wuhan SAR data set | | | | Shuguang heterogeneous data set | | | |
| --- | --- | --- | --- | --- | --- | --- | --- | --- | --- | --- | --- | --- |
| | FAR | MAR | OA | Kappa | FAR | MAR | OA | Kappa | FAR | MAR | OA | Kappa |
| 2 | **1.26** | 33.30 | 94.69 | 72.94 | **2.62** | 16.62 | 95.18 | 81.63 | 0.40 | 18.81 | 98.75 | 85.01 |
| 3 | 1.86 | **24.97** | **95.21** | **76.98** | 3.05 | **13.09** | **95.37** | **82.77** | 0.54 | **12.21** | **98.92** | **87.63** |
| 4 | 1.89 | 28.87 | 94.54 | 73.42 | 2.97 | 15.68 | 95.03 | 81.28 | **0.25** | 27.89 | 98.48 | 80.59 |
| 5 | 1.55 | 32.28 | 94.56 | 72.33 | 3.36 | 16.85 | 94.51 | 79.41 | 0.56 | 21.57 | 98.49 | 81.75 |

From Table V, we can see that the three-layer MSGCN achieves the best performances in terms of MAR, OA and Kappa on no matter which data set, which is consistent with the visual Interpretations. It can be possibly explained as follow: when the depth of network is 2, as the network is relatively shallow, the represented capability of features of nodes has not achieved the peak. When the number is 4 or 5, excessive smoothing may happen, as the GCN is intrinsic a low pass filter. In fact, the issue of excessive smoothing is a common limitation of the multi-layer GCN framework.

## V. CONCLUSION

A semisupervised change detection method based on graph convolutional network and the multiscale object-oriented analysis has been proposed in this paper to better address the CD tasks for both homogeneous and heterogeneous remote sensing images. To exploit the multiscale spatial information in high resolution images, the input image pair is firstly segmented through FNEA with different scale parameters, respectively, to obtain multiscale parcels (namely objects). Treating each parcel as one node, a graph construction strategy is used to form the parcel into a graph representation for each scale. Instead of concatenating hand-crafted features directly, we adopt a pre-trained U-net to extract pixel-wise high level features which are combined with the segmented results to obtain features of nodes in the graph representations. As several open RS datasets with labels for CD tasks have been available, no extra labeling workloads are needed. A novel multiscale graph convolutional network with each channel corresponding to one scale is proposed based on the property that the iterative training process helps to propagate the label information from labeled nodes to unlabeled ones, which allows us to use a fraction of labeled nodes to infer CD information for the whole image pair. Further, a fusion strategy is designed to incorporate the multiscale information. Three types of data, including openly optical data with thousands of high resolution image pairs, two pairs of high resolution SAR images, two pairs of heterogeneous images, are used to verify the effectiveness and superiority of the proposed method. The experimental results have demonstrated the superiority against some popular methods. Our future work is to extend the supervised framework to an unsupervised scenario and explore the possibility of distinguishing different kinds of change.

## ACKNOWLEDGMENT

The authors would like to thank Doc. Qingjie Liu of Beihang University for providing the free Beijing & Tianjin data set.